\documentclass{llncs}
\usepackage{llncsdoc}
\usepackage{array}
\usepackage[fleqn]{amsmath}
\usepackage{graphicx}

\usepackage[boxed]{algorithm2e}
\usepackage{multirow}
\usepackage{rotating}
\usepackage{booktabs}
\usepackage{longtable}
\usepackage{textcomp}
\usepackage{amssymb}
\usepackage{lmodern}
\usepackage[T1]{fontenc}
\usepackage{tipa}
\usepackage{caption}
\usepackage{subfig}
\newcommand{\PreserveBackslash}[1]{\let\temp=\\#1\let\\=\temp}
\newcolumntype{C}[1]{>{\PreserveBackslash\centering}p{#1}}
\newcolumntype{R}[1]{>{\PreserveBackslash\raggedleft}p{#1}}
\newcolumntype{L}[1]{>{\PreserveBackslash\raggedright}p{#1}}

\begin{document}
\title{Diversified Top-$k$ Partial MaxSAT Solving}
\author{Junping Zhou\inst{1} \and Huanyao Sun\inst{1} \and Feifei Ma\inst{2} \and Jian Gao\inst{1} \and Ke Xu\inst{3}  \and \\Minghao Yin\inst{1}}
\institute{
College of computer science and information technology, Northeast Normal University, Changchun 130117, China
\and
Institute of Software, Chinese Academy of Sciences,
Beijing 100080, China
\and
Department of Computer Science and Engineering, Beijing University of Aeronautics and Astronautics,
Beijing 100083, China.
}
\maketitle
\markboth{Diversified Top-$k$ Partial MaxSAT Solving}{Diversified Top-$k$ Partial MaxSAT Solving}

\begin{abstract}
We introduce a diversified top-$k$ partial MaxSAT problem, a combination of partial MaxSAT problem and enumeration problem. Given a partial MaxSAT formula $F$ and a positive integer $k$, the diversified top-$k$ partial MaxSAT is to find $k$ maximal solutions for $F$ such that the $k$ maximal solutions satisfy the maximum number of soft clauses of $F$. This problem can be widely used in many applications including community detection, sensor place, motif discovery, and combinatorial testing. We propose an approach for solving the problem. The concrete idea of the approach is to design an encoding EE which reduces diversified top-$k$ partial MaxSAT problem into partial MaxSAT problem, and then solve the resulting problem with state-of-art solvers. In addition, we present an algorithm MEMKC exactly solving the diversified top-$k$ partial MaxSAT. Through several experiments we show that our approach can be successfully applied to the interesting problem.\\

\end{abstract}
\section{Introduction}
Given a partial MaxSAT formula $F$ and a positive integer $k$, the diversified top-$k$ partial MaxSAT is a problem of finding $k$ maximal solutions for $F$ such that the $k$ maximal solutions satisfy the maximum number of soft clauses of $F$. As a matter of fact, it is a combination of partial MaxSAT problem and enumeration problem, which covers the traits of the two problems \cite{Chu 2017,Chu-Min2016,Michele2013,ChuMin2016}. For example, both the diversified top-$k$ partial MaxSAT and partial MaxSAT problems require to seek the solution(s) to satisfy the maximum number of soft clauses. Both the diversified top-$k$ partial MaxSAT and enumeration problems ask for enumerating some or all solutions. On the other hand, the proposed problem has its own characteristics and merits. It enumerates only $k$ solutions instead of all because in some applications listing $k$ solutions that are large and informative are preferred for a user \cite{Bernard1979}. Hence, diversified top-$k$ partial MaxSAT can open up a wide range of applications. For instance, the social network community detection can be solved using diversified top-$k$ partial MaxSAT based algorithm, by finding top-$k$ diversified communities covering most number of nodes in the social network \cite{Kempe2003}. The sensor place problem for water pollution monitoring, which is to select limited number of sensors guaranteeing the detection is maximized, can be expressed as the diversified top-$k$ partial MaxSAT problem \cite{Krause2007}. The diversified top-$k$ clique problem, used in motif discovery in molecular biology, can be also translated into diversified top-$k$ partial MaxSAT problem \cite{Zheng2011}. Additionally, the diversified top-$k$ partial MaxSAT problem can serve to detect the faults in complex software systems \cite{Yilmaz2006}. In this sense, the diversified top-k partial MaxSAT problem is versatile and valuable.

In this paper, we focus on exploring the diversified top-$k$ partial MaxSAT problem solving. The concrete proposal is to devise an efficient encoding from diversified top-$k$ partial MaxSAT problem into a target problem, and then solve the resulting problem with state-of-art solvers. Since the diversified top-$k$ partial MaxSAT problem is a generalization of the partial MaxSAT problem in some sense, we propose an encoding EE which reduces diversified top-$k$ partial MaxSAT into partial MaxSAT. Moreover, we present another algorithm MEMKC to exactly solve the diversified top-$k$ partial MaxSAT.The fundamental principle of the algorithm is to decompose the diversified top-$k$ partial MaxSAT into two stages, each of which corresponds to one problem: the first stage corresponds to the model enumeration problem, and the second corresponds to the Max-$k$-Cover problem. In the first stage, the algorithm enumerates each truth assignment that satisfies all hard clauses of a given partial MaxSAT instance $F$. In the second stage, it selects $k$ truth assignments from the results of the first stage to satisfy the maximum number of soft clauses of $F$. After that, we conduct experiments on diversified top-$k$ partial MaxSAT instances, which reveal that EE encoding is correct, effect, and efficient. The empirical investigation also shows that some application problems solved by converting them to diversified top-$k$ partial MaxSAT problem can obtain a better solution. For example, when we handle the diversified top-$k$ clique problem in dense graphs or hard graphs, it is more efficient to solve by reducing them to diversified top-$k$ partial MaxSAT problems than state-of-art diversified top-$k$ clique solvers.

The outline of the paper is as follows. We first present some related definitions in Section 2. In Section 3, we address the reduction of diversified top-$k$ partial MaxSAT into partial MaxSAT. We then propose another exact algorithm MEMKC to solve diversified top-$k$ partial MaxSAT in Section 4. In Section 5, two applications of the diversified top-$k$ partial MaxSAT are introduced. In Section 6, we show the experimental results. Finally, we conclude this paper.
\section{Preliminaries}
In this section, we will first present some definitions related this paper. A literal is either a Boolean variable (variable for short) $x$ or its negation $\neg x$. A positive literal is just a variable $x$, and the negative literal is the negation of a variable, i.e., $\neg x$. A clause is a disjunction of literals. The length of a clause is the number of literals that the clause contains. A clause is a unit clause if the length of the clause is 1. A formula $F$ in Conjunctive Normal Form (CNF) is a conjunction of clauses. The length of a formula $F$ expressed by $\lvert F \rvert$ is the number of clauses included in the formula. Any variable in $F$ can take a value {\it true} or {\it false}. A truth assignment for $F$ is a map that assigns each variable a value. Given a truth assignment, a clause is satisfied {\it iff} at least one literal in it take the value  {\it true} and a formula $F$ is satisfied {\it iff} each clause in $F$ is satisfied.

A partial MaxSAT formula in CNF is also a conjunction of clauses and the clauses can be distinguished into hard clauses and soft clauses. The hard clauses are un-relaxed constraints, all of which must be satisfied; while soft clauses are relaxed constrains, some of which can be unsatisfied. Thus, the partial MaxSAT formula can be expressed in a set of hard clauses and a set of soft clauses. Given a partial MaxSAT formula $F$ with a set of hard clauses $S_{h}$ and a set of soft clauses $S_{s}$, $S_{h}\cup S^{'}_{s}$ ($S^{'}_{s}\subseteq S_{s}$) is a maximal satisfiable subset of $F$ {\it iff} $S_{h}\cup S^{'}_{s}$ is satisfied, but addition of any clause in $S_{s}-S_{s}^{'}$ results in unsatisfied. We define the truth assignment satisfying any a maximal satisfiable subset is a maximal solution for a partial MaxSAT formula $F$. Given a partial MaxSAT formula $F$, the partial MaxSAT problem is to find a truth assignment for $F$ which satisfies not only all hard clauses but also the maximum number of soft clauses, and we call this truth assignment solution. Apparently, the truth assignment satisfying the largest maximal satisfiable subset is a solution of the partial MaxSAT problem.

\medskip\noindent
{\bfseries Definition 1.} Given a partial MaxSAT formula $F$ and a positive integer $k$, the diversified top-$k$ partial MaxSAT problem is to compute a set $R$, such that each $r_{i}\in R$ is a maximal solution of $F$, $\lvert R \rvert \leq k$, and the set $R$ can satisfy the maximum number of distinct soft clauses of $F$.

We define the set $R$ the solution of the diversified top-$k$ partial MaxSAT problem.

\section{Encoding}
In this section, we propose an encoding, named Expanding Encoding (EE), from diversified top-$k$ partial MaxSAT problem into partial MaxSAT problem. The central idea of EE is derived from the following observation. First of all, given a partial MaxSAT formula $F$ and a positive integer $k$, the diversified top-$k$ partial MaxSAT problem requires to find $k$ maximal solutions---that is, each variable in $F$ is set $k$ values though many of the values are repeated. So, we expand each variable into $k$ variables, i.e., $x_i$ is developed into $x_{i1}, x_{i2},..., x_{ik}$. Secondly, diversified top-$k$ partial MaxSAT problem and partial MaxSAT problem have something in common. They are both to compute a solution to satisfy the maximum number of soft clauses of $F$. Thus, we encode diversified top-$k$ partial MaxSAT problem into partial MaxSAT problem. The specific encoding process is addressed as the follows.

Given a positive integer $k$ and a partial MaxSAT formula $F$ with variables $x_1, x_2,..., x_n$, we create the partial MaxSAT formula as follows.
\begin {enumerate}
\item For each variable $x_i$ in $F$, we extend $x_i$ to $x_{i1}, x_{i2},..., x_{ik}$.
\item For each hard clause $\Box$$x_{a}\vee$ $\Box$$x_{b}\vee$ ... $\vee$  $\Box$$x_{p}$ in $F$, we develop the clause into $k$ hard clauses. They are (1) $\Box$$x_{a1} \vee$ $\Box$$x_{b1}\vee$...$\vee$ $\Box$$x_{p1}$, (2) $\Box$$x_{a2} \vee$ $\Box$$x_{b2}\vee...\vee$ $\Box$$x_{p2}$,... , ($k$) $\Box$$x_{ak} \vee$ $\Box$$x_{bk}\vee...\vee$ $\Box$$x_{pk}$, where "$\Box$" is either null or "\textlnot", and the corresponding symbol "$\Box$" in developed clauses is equal to the original hard clause.
\item For each soft clause  $\Box$$x_{q}\vee$ $\Box$$x_{t}\vee...\vee$  $\Box$$x_{z}$ in $F$, we expand the clause into a clause with $k$ times length as the original one. That is ($\Box$$x_{q1}\vee$ $\Box$$x_{t1}\vee...\vee$ $\Box$$x_{z1}$)$\vee$($\Box$$x_{q2}\vee$ $\Box$$x_{t2}\vee...\vee$ $\Box$$x_{z2})\vee ... \vee$ ($\Box$$x_{qk}\vee$ $\Box$$x_{tk}\vee...\vee$ $\Box$$x_{zk})$, where "$\Box$" is either null or "\textlnot", and the corresponding symbol "$\Box$" in developed clauses is also equal to the original soft clause.
\end{enumerate}

Since hard clauses are un-relaxed constraints and the values which are set to variables in $F$ must satisfy each original hard clause, the auxiliary hard clauses are added into the converted partial MaxSAT formula. In addition, because soft clauses are relaxed constrains and can be satisfied if just one literal set to {\it true}, the original soft clause grows longer in the converted partial MaxSAT formula. Generally speaking, if the original diversified top-$k$ partial MaxSAT has $n$ variables, $m$ hard clauses, and $m^{'}$ soft clauses, the partial MaxSAT contains $k\times n$ variables, $k\times m$ hard clauses, and $m^{'}$ soft clauses.

\medskip\noindent
{\bfseries Example 1.} Let $F=S_h \cup S_s$ be a partial MaxSAT formula, where $S_h=\{x_1\vee x_2, \neg x_1\vee \neg x_2\}$ is the set of hard clauses, and $S_s=\{x_1, x_2\}$ is the set of soft clauses. When $k=2$, the partial MaxSAT encoding of the diversified top-$k$ partial MaxSAT instance $F$ is formed by the hard clauses $c_1= x_{11} \vee x_{21}, c_2= x_{12}\vee x_{22}, c_3=\neg x_{11}\vee \neg x_{21}, c_4=\neg x_{12} \vee \neg x_{22}$, and the soft clauses $c_5=x_{11}\vee x_{12}$, $c_6=x_{21}\vee x_{22}$.

\section{Reducing Diversified top-$k$ Partial MaxSAT Problem} 
In this section, we define a method, named MEMKC, to exactly solve the diversified top-$k$ partial MaxSAT problem. The method split the original problem into model enumeration \cite{Morgado2005} and Max-$k$-Cover \cite{Chang2013}. This method MEMKC firstly enumerates every truth assignment satisfying all hard clauses of a given partial MaxSAT formula $F$, and such truth assignment is called model. Let the soft clauses of the given partial MaxSAT formula $F$ be a universe set $U$, where each soft clause is an element of $U$, and let the soft clauses satisfied by a model be a subset of U. Obviously, to solve a top-k Partial MaxSAT problem can be reduced to a Max-k-cover problem, i.e. seeking out $k$ subset from the subsets returned by previous model enumeration process to cover the maximum number of elements in the universe set.

In the following, we will describe the framework of MEMKC algorithm in Fig.\ref{fig1}. This algorithm includes the ME (Model Enumeration) and MKC (Max-$k$-Cover) functions, and takes the partial MaxSAT formula $F$ and a positive integer $k$ as the inputs. In the algorithm, $S_h$ (resp. $S_s$) denotes the formula only containing hard clauses (resp. soft clauses), which is obtained by deleting all the soft clauses (resp. hard clauses) from $F$. $SM$ is a set storing the selected models. At first, the algorithm invokes the ME to enumerate all models stored in the set $S$. Then, the MKC is called to select $k$ models satisfying the maximum number of soft clauses. Fig.\ref{fig2} and Fig.\ref{fig3} propose the ME and MKC functions respectively.  In the two functions, $s$, $s_1$, and $s_2$ is the current models expressed in ${x_i=value, x_j=value,...}$, where $x_i$, $x_j$, ... are the variables, and $value$ is either $true$ or $false$. The symbol "$\Box$" is either null or "\textlnot". When the symbol "$\Box$" in $\Box$$x=true$ is null, the value of the variable $x$ is $true$; otherwise $false$. $F\lvert_{x=true}$ (resp. $F\lvert_{x=false}$) is the result of applying a simplified rule, i.e., removing all clauses containing the literal $x$ (resp.$\neg x$) from $F$, and deleting all occurrences of $\neg x$ (resp.$x$) from the other clauses in $F$. $S/s$ is a set of models obtained by deleting the model $s$ from $S$. $F\lvert_{s=true}$ is a formula received by dropping the satisfied clauses by the model $s$ from \textit {F}. $SN_i$ (\textit {i}=1 or 2) is a variable recoding the number of soft clauses satisfied by the models in $SM$.
\begin{figure}
\setlength{\abovecaptionskip}{0.cm}
 \setlength{\belowcaptionskip}{-1.cm}
\begin{algorithm}[H]
  {\bfseries Algorithm} MEMKC(\em{F},$k$)\\
$S=\emptyset$, $SM=\emptyset$; \qquad  $// \emptyset$ is an empty set.\\
$S$=ME($S_h$, $\emptyset$)\;
{\bfseries return} MKC($S_s$, $S$, $k$, SM).
\end{algorithm}
\caption[]{The MEMKC algorithm}
\label{fig1}
\end{figure}
\begin{figure}[htbp]
\centering

\centering
\begin{minipage}[t]{0.49\textwidth}
\centering

\begin{algorithm}[H]
 {\bfseries Function}  ME ($F$,$s$)\\

   {\bfseries while} {$F$ contains a unit clause $\Box$$x$}\\
  { \quad $F$ =$F$$\lvert_{\Box x=true}$\;
    \quad add $\Box$$x=true$ to $s$ \;}

 {\bfseries if} {$F$ has an empty clause}\\
  {{\bfseries then return} $\emptyset$;}

  {\bfseries if} {$F$ is empty}  \\
  {{\bfseries then return} $\{s\}$;}

 $s_1=s;~s_2=s$\;
  pick a variable $x$ from $F$ \;
  $F_1$ =$F$$\lvert_{x=true}$\;
  add ${x =true}$ to $s_1$ \;

  $F_2$ =$F\lvert_{x=false}$\;
  add ${x =false}$ to $s_2$ \;

  {\bfseries return} ME($F_1$,$s_1$) $\cup$ ME($F_2$,$s_2$).
\end{algorithm}

\caption[]{The ME function}
\label{fig2}

\end{minipage}
\begin{minipage}[t]{0.49\textwidth}
\centering
\begin{algorithm}[H]
  {\bfseries Function} MKC ($F$,$S$,$k$,$SM$)

  {\bfseries if} {$k$=0 or $S=\emptyset$}  \\
  {{\bfseries then} {\bfseries return} 0\;}

  pick a model $s$ from $S$\;
   $S=S/s$ \;
   $F_1$ = $F$\;
  $F_2$ = $F$$\lvert_{s=true}$\;
   $SN_1$=MKC($F_1$, $S$, $k$, $SM$)\;
   $SN_2$=MKC($F_2$,\textit {S}, \textit {k}-1,$SM\cup \{s\}$)+ $(\lvert F\rvert - \lvert F_{2}\rvert)$\;
  {\bfseries return} $max(SN_1, SN_2)$. \\

\end{algorithm}
\caption[]{The MKC function}
\label{fig3}
\end{minipage}
\end{figure}

\section{Applications}
In this section, we will discuss the applications of the diversified top-$k$ partial Max-SAT problem.
\subsection{Diversified Top-$k$ Clique}
The problem of diversified top-$k$ clique is to pick up $k$ maximal cliques which cover the maximum number of vertices \cite{Long2015}. This problem has wide real-world applications including social networks, complex networks, and molecular biology \cite{N. Berry 2004,C. Lee 2010,X. Zheng 2011}. In addition, diversified top-$k$ clique problem can be solved by converting into diversified top-$k$ partial MaxSAT. Suppose \emph G=<\emph V,\emph E> is an undirected and simple graph, where \emph V=$\{v_1, v_2,..., v_n \}$ is a set of vertices, \emph E=$\{ e_1, e_2,..., e_m \}$ is a set of edges, and each $e_i$ is expressed as a 2-tuple <$v_t$, $v_j$> ($1\leq t,j\leq n, t\neq j$). Given a positive integer $k$, the encoding from diversified top-$k$ clique into diversified top-$k$ partial MaxSAT is as follows. The integer $k$ remains the same. The hard clauses are composed of all the non-adjacent vertices in $G$, i.e., $\neg v_i \vee \neg v_j$, where < $v_i$ , $v_j$ > $\notin E$, and the soft part is the unit clauses formed by every vertex in $G$, i.e., $v_i$, where $v_i \in V$.

\medskip\noindent
{\bfseries Example 2.} Given a positive integer $k$, an undirected and simple graph \emph G=<\emph V, \emph E>, where \emph V=$\{v_1, v_2, v_3, v_4\}$ and \emph E=\{<$v_1, v_2$>, <$v_2, v_3$>, <$v_3, v_4$>, <$v_1, v_4$>\}, the diversified top-\emph k partial MaxSAT encoding of the diversified top-$k$ clique in $G$ consists of the integer $k$, hard clauses $\neg v_1 \vee \neg v_3$, $\neg v_2 \vee \neg v_4$, and soft clauses $v_1$, $v_2$, $v_3$, $v_4$.

\subsection{Diversified Top-$k$ Covering Arrays}

Covering Arrays (CAs) are interesting objects of study in combinatorics. It plays an important role in factorial designs in which each treatment is a combination of factors at different levels. A CA of run size $N$, factor number $M$, strength $t$ can be denoted by CA$(N,s_1,s_2,\ldots,s_M,t)$. It is an $N\times M$ matrix satisfying the following constraints:
\begin{enumerate}
\item There are exactly $s_i$ symbols appearing in each column \textit {i} $(1\le i\le M)$.
\item In every $N\times t$ sub-array, each ordered combination of symbols from the $t$ columns appears at least once in rows.
\end{enumerate}
$s_i$ is refered to as the level of factor $i$. By combining equal entries in $s_i$, a CA$(N,s_1,s_2,\ldots,s_M,t)$ is represented in the shortened form CA$(N,{s_{i_1}}^{a_1},{s_{i_2}}^{a_2}$, $\ldots,t)$, where $a_1$, $a_2$, \ldots indicate the number of factors at level $s_{i_1}$, $s_{i_2}$, \ldots.

For many complex software systems, there are usually different components$/$ parameters, each of which can take a number of values. System faults usually arise from the interaction of several parameters$/$components. Combinatorial testing \cite{lin_2015,kuhn_2002,AETG_TSE97,KWG04,Nie_CSvy11} is an important black-box testing technique to reveal such faults. CAs have been used as test suites in Combinatorial testing for more than two decades. Each row of the CA represents a test case, and the $i$-th column corresponds to the value of parameter $p_i$ in each test case. For any $t$ columns of the array, the $N\times t$ sub-array covers all value combinations of the corresponding $t$ parameters. Therefore, a CA of strength $t$ can be used as a test suite to detect the faults caused by the interactions of $t$ parameters.

Ideally, one would like to examine the system's behavior under all $t$-way combinations of the parameter values. But this kind of exhaustive testing may be very costly for a non-trivial software system. In some occasions, resource or time only permits to execute a certain number of test cases. It is highly desirable to maximize the coverage of these test cases. Given parameters, coverage criteria, and a positive integer $k$, a diversified top-$k$ CA is a $k\times M$ sub-matrix of a CA in which the number of ordered combination of $t$ values from different columns is maximized. The diversified top-$k$ CA problem can be also solved by transforming it to diversified top-$k$ partial MaxSAT.
Suppose there is a positive integer $k$ and a CA$(N,s_1,s_2,\ldots,s_M,t)$.Then each value combination can be modeled as a variable. If two value combinations contradict each other, then we add a hard clause, i.e., $\neg var_i \vee \neg var_j$, where the value combination modeled as $var_i$ contradicts with $var_j$. In addition, the soft part is formed by each variable, i.e., $var_i$. The integer $k$ is unchanged.

\medskip\noindent
{\bfseries Example 3.} Given a positive integer $k$ and a CA$(N,{s_{i_1}}^{a_1},{s_{i_2}}^{a_2}, {s_{i_3}}^{a_3},t)$, where $N=2$, $M=3$, $t=2$, $s_{i_1}=1$, $s_{i_2}=1$, $s_{i_3}=2$, the diversified top-$k$ partial MaxSAT encoding of the diversified top-$k$ CA consists of the integer $k$, the hard clauses (1)$\neg var_2 \vee \neg var_3$, (2)$\neg var_4 \vee \neg var_5$, and the soft clauses $(1)var_1, (2)var_2,..., (5)var_{5}$.

Note that, in general, $s_{i_j}$>1. This example set that way is only to illustrate how the transformation works.

\section{Experimental Results}
In this section, we perform experiments to evaluate the effectiveness and efficiency of EE encoding from the comparison of our algorithm MEMKC, diversified top-$k$ clique solver EnumKOpt \cite{Long2015}, diversified top-$k$ CA solver $k$-CA \cite{lin_2015}, and the state-of-the-art partial MaxSAT solvers Openwbo \cite{Ruben2014}, Dist \cite{Shaowei2014}, and CCLS \cite{Luo2015}.
\subsection{Experimental Preliminaries}
We implement MEMKC algorithm in C++ programming language. Since enumerating all models is computationally intractable, MEMKC is only used in Part 1 to evaluate the small-scale instances. EnumKOpt is the best current approximate solver for diversified top-$k$ clique. Diversified top-$k$ CA solver $k$-CA implemented in C++ is an approximate solver by greedily calling the state-of-the-art CA solver TCA \cite {lin_2015}. Dist and CCLS are excellent approximate solvers for partial MaxSAT. The time limits of EnumKOpt, $k$-CA, Dist, and CCLS are set to 1800, 300, 300, 300 seconds respectively when they execute once on each instance. Openwbo, an exact solver, wins the first prize in unweighted partial MaxSAT track in MaxSAT Evaluation 2016. The cut off time of Openwbo is set to 30000 seconds. In the experimental results, if a solver fails to find a feasible solution in the cut off time, the corresponding results are marked with "-". The execution time retured is in seconds. \#uncov represents the number of unsatisfied soft clauses (uncovered nodes or value combinations) solved by solvers. And the results obtained by Dist and CCLS are the average values by executing ten times on each instance. All experiments are running on a workstation under Linux with 8 cores of Inter(R) Xeon(R) E7-4820 v2 @2.00GHz CPU and 8GB RAM.

\begin{figure}[!htb]
\setlength{\abovecaptionskip}{10pt}
 \setlength{\belowcaptionskip}{-20pt}
\centering
\subfloat{\includegraphics[width = 0.49\linewidth,height=3.5cm]{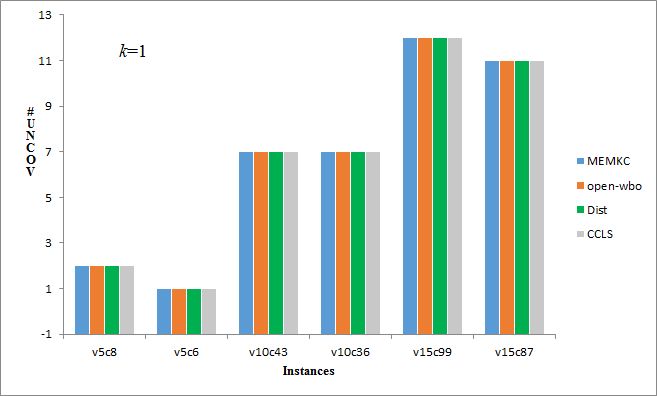}}\hfill
\subfloat{\includegraphics[width = 0.49\linewidth,height=3.5cm]{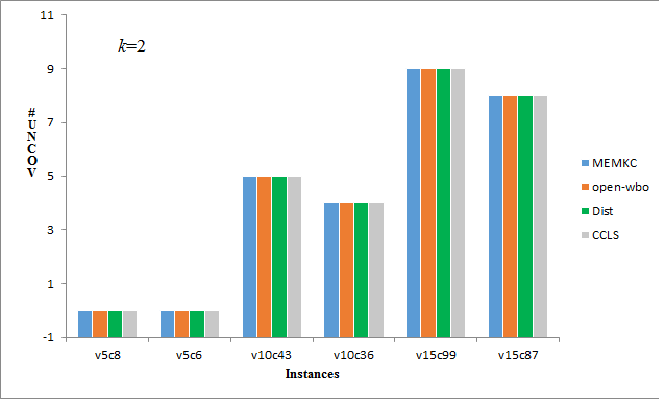}}\hfill
\subfloat{\includegraphics[width = 0.49\linewidth,height=3.5cm]{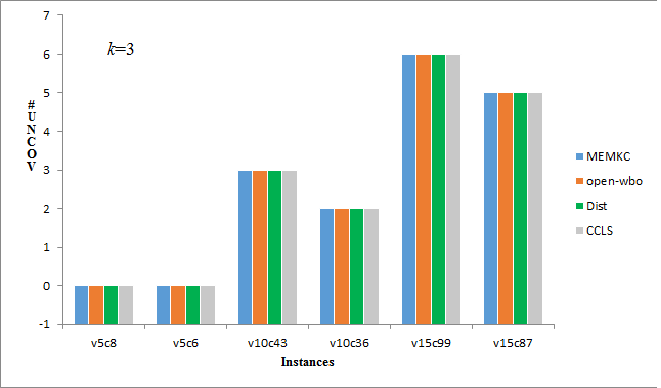}}\hfill
\subfloat{\includegraphics[width = 0.49\linewidth,height=3.5cm]{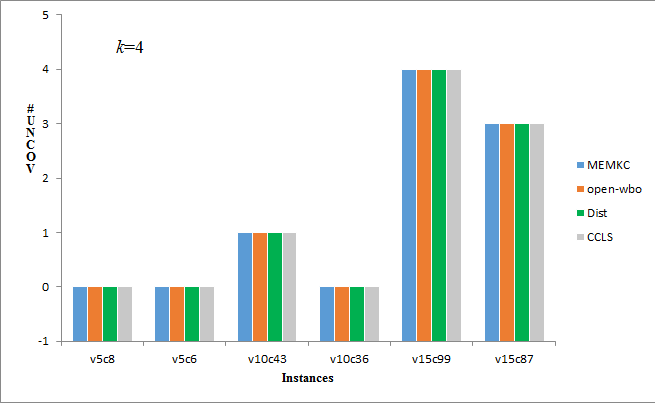}}\hfill
\subfloat{\includegraphics[width = 0.49\linewidth,height=3.5cm]{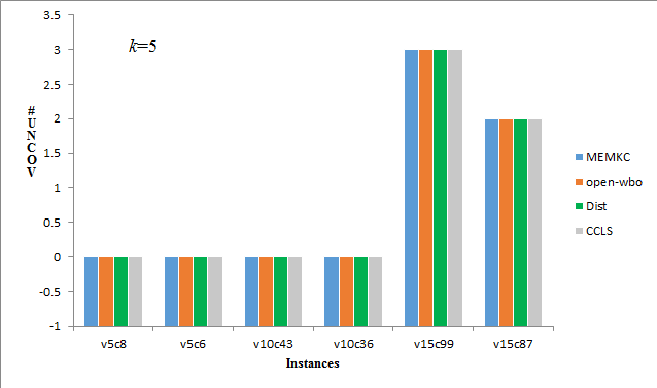}}\hfill
\subfloat{\includegraphics[width = 0.49\linewidth,height=3.5cm]{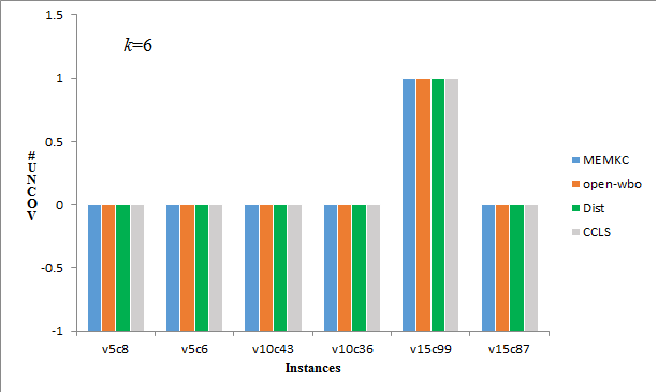}}\\
\caption[]{Experiment results on random instances varying $k$ from 1 to 6}
\label{fig4}
\end{figure}

\subsection{Experimental Results}

~~~{\bfseries Part 1:} This experiment is to demonstrate the effectiveness of MEMKC. It should be noted that since MEMKC need to enumerate all the Fig.\ref{fig4} illustrates the number of unsatisfied soft clauses returned by MEMKC solver and partial MaxSAT solvers, including exact solver Openwbo and approximate solvers Dist, CCLS, on random diversified top-$k$ partial MaxSAT instances. These random instances solved by the partial MaxSAT solvers are encoded with EE encoding. In the experiment, $vicj$ represents the instance with $i$ variables, $j-i$ hard clauses, and $i$ soft clauses. The results reported in Fig.\ref{fig4} suggest that MEMKC is effective because each solver can obtain the same results on the identical instances.

{\bfseries Part 2:} The purpose of the experiment is to verify the effectiveness of the EE encoding. Table 1 presents the performance of three partial MaxSAT solvers on instances encoded from random diversified top-$k$ partial MaxSAT instances with EE encoding. In this experiment, we vary $k$ from 1 to 6. The scale of these instances (expressed by var\textit {i}) ranges from 40 to 100 variables and 200 to 1200 clauses, where the number of soft clauses is from 40 to 100. Table 1 records the mean number of unsatisfied soft clauses for each solver in a set of 30 instances. By analyzing the results in Table 1, we can see that the instances with a few variables can be exact solved, and all instances can be successfully settled by approximate solvers, Dist and CCLS, which reveals that the EE encoding is effect.

\begin{table}[htbp]
  \setlength{\abovecaptionskip}{-2.pt}
 \setlength{\belowcaptionskip}{-5.pt}
  \scriptsize
  \centering
  \caption{The comparison results of three partial MaxSAT solvers on random instances }
    \newsavebox{\tablebox}
    \begin{lrbox}{\tablebox}
    \begin{tabular}{C{0.7cm}C{0.7cm}R{2cm}R{2cm}R{2cm}}
    \toprule
    \multicolumn{1}{r}{\textbf{Instance}} & \textit{\textbf{k}} & \textbf{Openwbo} & \textbf{Dist } & \textbf{CCLS} \\
    \midrule
    \multirow{6}[2]{*}{var40} & 1     & 26.77(30) & 26.77(30) & 26.77(30) \\
          & 2     & 16.10(30) & 16.10(30) & 16.10(30) \\
          & 3     & 7.47(30) & 7.47(30) & 7.47(30) \\
          & 4     & 1.37(30) & 1.37(30) & 1.37(30) \\
          & 5     & 0.00(30) & 0.00(30) & 0.00(30) \\
          & 6     & 0.00(30) & 0.00(30) & 0.00(30) \\
    \midrule
    \multirow{6}[2]{*}{var50} & 1     & 35.57(30) & 35.57(30) & 35.57(30) \\
          & 2     & 23.13(30) & 23.13(30) & 23.13(30) \\
          & 3     & 12.73(30) & 12.73(30) & 12.73(30) \\
          & 4     & 4.43(30) & 4.43(30) & 4.43(30) \\
          & 5     & 0.07(30) & 0.07(30) & 0.07(30) \\
          & 6     & 0.00(30) & 0.00(30) & 0.00(30) \\
    \midrule
    \multirow{6}[2]{*}{var60} & 1     & 44.57(30) & 44.57(30) & 44.57(30) \\
          & 2     & 30.63(30) & 30.63(30) & 30.63(30) \\
          & 3     & 18.67(30) & 18.67(30) & 18.67(30) \\
          & 4     & 8.80(30) & 8.80(30) & 8.80(30) \\
          & 5     & 1.53(30) & 1.53(30) & 1.53(30) \\
          & 6     & 0.00(30) & 0.00(30) & 0.00(30) \\
    \midrule
    \multirow{6}[2]{*}{var70} & 1     & -     & 53.30(30) & 53.30(30) \\
          & 2     & -     & 38.33(30) & 38.33(30) \\
          & 3     & -     & 24.83(30) & 24.83(30) \\
          & 4     & -     & 13.50(30) & 13.50(30) \\
          & 5     & -     & 4.60(30) & 4.60(30) \\
          & 6     & -     & 0.00(30) & 0.00(30) \\
    \midrule
    \multirow{6}[2]{*}{var80} & 1     & -     & 62.47(30) & 62.47(30) \\
          & 2     & -     & 46.33(30) & 46.33(30) \\
          & 3     & -     & 31.90(30) & 31.90(30) \\
          & 4     & -     & 19.33(30) & 19.33(30) \\
          & 5     & -     & 8.80(30) & 8.80(30) \\
          & 6     & -     & 1.10(30) & 1.10(30) \\
    \midrule
    \multirow{6}[2]{*}{var90} & 1     & -     & 71.60(30) & 71.60(30) \\
          & 2     & -     & 54.57(30) & 54.57(30) \\
          & 3     & -     & 39.20(30) & 39.20(30) \\
          & 4     & -     & 25.43(30) & 25.43(30) \\
          & 5     & -     & 13.33(30) & 13.33(30) \\
          & 6     & -     & 3.90(30) & 3.90(30) \\
    \midrule
    \multirow{6}[2]{*}{var100} & 1     & -     & 80.73(30) & 80.73(30) \\
          & 2     & -     & 62.97(30) & 62.97(30) \\
          & 3     & -     & 46.73(30) & 46.73(30) \\
          & 4     & -     & 31.80(30) & 31.80(30) \\
          & 5     & -     & 16.67(30) & 16.67(30) \\
          & 6     & -     & 7.90(30) & 7.90(30) \\
    \bottomrule
    \end{tabular}%
    \end{lrbox}
    \scalebox{1.0}{\usebox{\tablebox}}
  \label{tab:addlabel}%
\end{table}%

{\bfseries Part 3:} We carry out three experiments comparing with the diversified top-$k$ clique solver EnumKOpt and three partial MaxSAT solvers on graphs to illustrate the efficiency of EE encoding. These instances solved by the partial MaxSAT solvers are first encoded from graphs into diversified top-$k$ partial MaxSAT problem and then converted into partial MaxSAT problem with EE encoding. In the first experiment, we randomly generate graphs with 40 to 100 vertices and 10\% probability that two vertices have an edge. In the second experiment, the generated random graphs vary probabilities of an edge existing between two vertices from 10\% to 70\%, and the number of vertices in all graphs is 60. In the results of the experiment, each graph is represented by Pj, where \textit {j} is the probability. In the last experiment, we choose the well-known BHOSLIB benchmark, which is famous for its hardness \cite{KeXu 2006}.

Table 2 presents the results of the first experiment, which compares the diversified top-$k$ clique solver with partial MaxSAT solvers on random sparse graphs by varying the number of vertices. We use Vi to express each instance, where \textit {i} is the number of vertices. As can be seen from Table 2, exact solver Openwbo can deal with the instances with small scale and all instances can be rapidly solved by the approximate solvers. In addition, comparing with all known optimal solutions found by Openwbo, almost all approximate solvers can obtain the optimal solutions. This indicates that EE encoding and EnumKOpt have the equal effectiveness when solving the diversified top-$k$ clique problem in sparse graphs. The results of the second experiment are showed in Table 3. This experiment compares the diversified top-$k$ clique solver with three partial MaxSAT solvers on random graphs with varying probabilities that two vertices have an edge. Among these results, we observe the performance of EnumKOpt becomes worse with the increasing probability, while the performance of approximate solvers for partial MaxSAT is stable. Moreover, the exact solver Openwbo for partial MaxSAT can find solutions on random graphs with high probabilities. This illustrates that it is more appropriate for diversified top-$k$ clique problem to be solved by converting to diversified top-$k$ partial MaxSAT problem when they handle the dense graphs. Table 4 summarizes the experimental results of EnumKOpt and three partial MaxSAT solvers on instances from BHOSLIB. We can find that these graphs cannot be solved by EnumKOpt and Openwho, but can be efficiently obtained the solutions by translating these graphs into diversified top-$k$ partial MaxSAT and further encoding to partial MaxSAT with EE encoding. This illustrates that EE encoding is efficient on hard instances and it will be also efficient when we solve the diversified top-$k$ cliques on hard instances by conventing it to diversified top-$k$ partial MaxSAT.

\begin{table}[htbp]
  \scriptsize
  \centering
  \caption{The comparison between EnumKOpt and three partial MaxSAT solvers on random graphs with varying the number of vertices}
    \begin{tabular}{cC{0.7cm}rrrrrrrr}
    \toprule
    \multicolumn{1}{r}{\multirow{2}[4]{*}{\textbf{Instance}}} & \multirow{2}[4]{*}{\textit{\textbf{k}}} & \multicolumn{2}{c}{\textbf{EnumKOpt}} & \multicolumn{2}{c}{\textbf{Openwbo}} & \multicolumn{2}{c}{\textbf{Dist}} & \multicolumn{2}{c}{\textbf{CCLS}} \\
\cmidrule{3-10}          &       & \textbf{\#uncov} & \textbf{Time} & \textbf{\#uncov} & \textbf{Time} & \textbf{\#uncov} & \textbf{Time} & \textbf{\#uncov} & \textbf{Time} \\
    \midrule
    \multirow{6}[2]{*}{V40} & 1     & 37    & 0     & 37    & 0.01  & 37    & 0.02  & 37    & 0 \\
          & 2     & 34    & 0     & 34    & 0.22  & 34    & 0.03  & 34    & 0 \\
          & 3     & 31    & 0     & 31    & 3.61  & 31    & 0.02  & 31    & 0 \\
          & 4     & 28    & 0     & 28    & 84.8  & 28    & 0.03  & 28    & 0 \\
          & 5     & 26    & 0     & 25    & 2069.67 & 25    & 0.01  & 25    & 0 \\
          & 6     & 24    & 0     &       & -     & 23    & 0.01  & 23    & 0 \\
    \midrule
    \multirow{6}[2]{*}{V50} & 1     & 47    & 0     & 47    & 0.02  & 47    & 0.01  & 47    & 0 \\
          & 2     & 44    & 0     & 44    & 0.97  & 44    & 0.01  & 44    & 0 \\
          & 3     & 41    & 0     & 41    & 23.64 & 41    & 0.01  & 41    & 0 \\
          & 4     & 38    & 0     & 38    & 1021.6 & 38    & 0.01  & 38    & 0 \\
          & 5     & 35    & 0     & -     & -     & 35    & 0.01  & 35    & 0 \\
          & 6     & 33    & 0     & -     & -     & 33    & 0.01  & 33    & 0 \\
    \midrule
    \multirow{6}[2]{*}{V60} & 1     & 57    & 0     & 57    & 0.04  & 57    & 0.01  & 57    & 0 \\
          & 2     & 54    & 0     & 54    & 2.83  & 54    & 0.01  & 54    & 0 \\
          & 3     & 51    & 0     & 51    & 334.47 & 51    & 0.01  & 51    & 0 \\
          & 4     & 48    & 0     & 48    & 29152.67 & 48    & 0.01  & 48    & 0 \\
          & 5     & 45    & 0     & -     & -     & 45    & 0.01  & 45    & 0 \\
          & 6     & 42    & 0     & -     & -     & 42    & 0.01  & 42    & 0 \\
    \midrule
    \multirow{6}[2]{*}{V70} & 1     & 67    & 0     & 67    & 0.06  & 67    & 0.01  & 67    & 0 \\
          & 2     & 64    & 0     & 64    & 5.28  & 64    & 0.01  & 64    & 0 \\
          & 3     & 61    & 0     & 61    & 811.58 & 61    & 0.01  & 61    & 0 \\
          & 4     & 58    & 0     & -     & -     & 58    & 0.01  & 58    & 0 \\
          & 5     & 55    & 0     & -     & -     & 55    & 0.01  & 55    & 0 \\
          & 6     & 52    & 0     & -     & -     & 52    & 0.02  & 52    & 0 \\
    \midrule
    \multirow{6}[2]{*}{V80} & 1     & 77    & 0     & 77    & 0.16  & 77    & 0.01  & 77    & 0 \\
          & 2     & 74    & 0     & 74    & 14.79 & 74    & 0.01  & 74    & 0 \\
          & 3     & 71    & 0     & 71    & 4672.18 & 71    & 0.01  & 71    & 0 \\
          & 4     & 68    & 0     & -     & -     & 68    & 0.01  & 68    & 0 \\
          & 5     & 65    & 0     & -     & -     & 65    & 0.02  & 65    & 0.01 \\
          & 6     & 62    & 0     & -     & -     & 62    & 0.02  & 62    & 0.01 \\
    \midrule
    \multirow{6}[2]{*}{V90} & 1     & 86    & 0     & 86    & 0.37  & 86    & 0.01  & 86    & 0 \\
          & 2     & 83    & 0     & 83    & 26.97 & 83    & 0.01  & 83    & 0 \\
          & 3     & 80    & 0     & -     & -     & 80    & 0.01  & 80    & 0 \\
          & 4     & 77    & 0     & -     & -     & 77    & 0.02  & 77    & 0.01 \\
          & 5     & 74    & 0     & -     & -     & 74    & 0.02  & 74    & 0.01 \\
          & 6     & 71    & 0     & -     & -     & 71    & 0.02  & 71    & 0.01 \\
    \midrule
    \multirow{6}[2]{*}{V100} & 1     & 96    & 0     & 96    & 0.29  & 96    & 0.01  & 96    & 0 \\
          & 2     & 93    & 0     & 93    & 53.55 & 93    & 0.01  & 93    & 0 \\
          & 3     & 90    & 0     & -     & -     & 90    & 0.02  & 90    & 0 \\
          & 4     & 87    & 0     & -     & -     & 87    & 0.02  & 87    & 0.01 \\
          & 5     & 84    & 0     & -     & -     & 84    & 0.02  & 84    & 0.01 \\
          & 6     & 81    & 0     & -     & -     & 81    & 0.03  & 81    & 0.01 \\
    \bottomrule
    \end{tabular}%
  \label{tab:addlabel}%
\end{table}%

\begin{table}[htbp]
  \scriptsize
  \centering
  \caption{The comparison between EnumKOpt and three partial MaxSAT solvers on random graphs by varying probabilities that two vertices have an edge}
    \begin{tabular}{cC{0.7cm}rrrrrrrr}
    \toprule
    \multicolumn{1}{r}{\multirow{2}[4]{*}{\textbf{Instance}}} & \multirow{2}[4]{*}{\textit{\textbf{k}}} & \multicolumn{2}{c}{\textbf{EnumKOpt}} & \multicolumn{2}{c}{\textbf{Openwbo}} & \multicolumn{2}{c}{\textbf{Dist}} & \multicolumn{2}{c}{\textbf{CCLS}} \\
\cmidrule{3-10}          &       & \textbf{\#uncov} & \textbf{Time} & \textbf{\#uncov} & \textbf{Time} & \textbf{\#uncov} & \textbf{Time} & \textbf{\#uncov} & \textbf{Time} \\
    \midrule
    \multirow{6}[2]{*}{P10} & 1     & 57    & 0     & 57    & 0.04  & 57    & 0.01  & 57    & 0 \\
          & 2     & 54    & 0     & 54    & 2.83  & 54    & 0.01  & 54    & 0 \\
          & 3     & 51    & 0     & 51    & 334.47 & 51    & 0.01  & 51    & 0 \\
          & 4     & 48    & 0     & 48    & 29152.67 & 48    & 0.01  & 48    & 0 \\
          & 5     & 45    & 0     & -     & -     & 45    & 0.01  & 45    & 0 \\
          & 6     & 42    & 0     & -     & -     & 42    & 0.01  & 42    & 0 \\
    \midrule
    \multirow{6}[2]{*}{P20} & 1     & 56    & 0     & 56    & 0.06  & 56    & 0.01  & 56    & 0 \\
          & 2     & 52    & 0     & 52    & 6.03  & 52    & 0.01  & 52    & 0 \\
          & 3     & 49    & 0     & 48    & 255.7 & 48    & 0.01  & 48    & 0 \\
          & 4     & 46    & 0     & -     & -     & 44    & 0.01  & 44    & 0 \\
          & 5     & 43    & 0     & -     & -     & 40    & 0.02  & 40    & 0 \\
          & 6     & 40    & 0     & --    & --    & 37    & 0.01  & 37    & 0 \\
    \midrule
    \multirow{6}[2]{*}{P30} & 1     & 55    & 0     & 55    & 0.1   & 55    & 0     & 55    & 0 \\
          & 2     & 50    & 0     & 50    & 10.71 & 50    & 0.01  & 50    & 0 \\
          & 3     & 46    & 0     & 45    & 2153.82 & 45    & 0.01  & 45    & 0 \\
          & 4     & 42    & 0     & -     & -     & 40    & 0.01  & 40    & 0 \\
          & 5     & -     & -     & -     & -     & 35    & 0.01  & 35    & 0 \\
          & 6     & -     & -     & -     & -     & 31    & 0.03  & 31    & 0 \\
    \midrule
    \multirow{6}[2]{*}{P40} & 1     & 54    & 0     & 54    & 0.08  & 54    & 0.01  & 54    & 0 \\
          & 2     & 49    & 0     & 48    & 15.43 & 48    & 0.01  & 48    & 0 \\
          & 3     & 44    & 0     & 42    & 6116.46 & 42    & 0.01  & 42    & 0 \\
          & 4     & -     & -     & -     & -     & 36    & 0.01  & 36    & 0 \\
          & 5     & -     & -     & -     & -     & 31    & 0.02  & 31    & 0 \\
          & 6     & -     & -     & -     & -     & 26    & 0.01  & 26    & 0 \\
    \midrule
    \multirow{6}[2]{*}{P50} & 1     & 52    & 0     & 52    & 0.11  & 52    & 0.01  & 52    & 0 \\
          & 2     & 46    & 0     & 44    & 11.67 & 44    & 0.01  & 44    & 0 \\
          & 3     & -     & -     & 38    & 6630.16 & 38    & 0.01  & 38    & 0 \\
          & 4     & -     & -     & -     & -     & 32    & 0.02  & 32    & 0 \\
          & 5     & -     & -     & -     & -     & 26    & 0.03  & 26    & 0.01 \\
          & 6     & -     & -     & -     & -     & 20    & 0.42  & 20    & 0.01 \\
    \midrule
    \multirow{6}[2]{*}{P60} & 1     & 45    & 0     & 45    & 0.36  & 45    & 0.04  & 45    & 0 \\
          & 2     & -     & -     & 30    & 12.56 & 30    & 0.06  & 30    & 0 \\
          & 3     & -     & -     & 18    & 1178.02 & 18    & 0.06  & 18    & 0 \\
          & 4     & -     & -     & 8     & 4947.78 & 8     & 0.05  & 8     & 0 \\
          & 5     & -     & -     & 1     & 1.81  & 1     & 0.01  & 1     & 0 \\
          & 6     & -     & -     & 0     & 0     & 0     & 0.01  & 0     & 0 \\
    \midrule
    \multirow{6}[2]{*}{P70} & 1     & 39    & 0     & 37    & 0.08  & 37    & 0.03  & 37    & 0 \\
          & 2     & -     & -     & 19    & 1.48  & 19    & 0.03  & 19    & 0 \\
          & 3     & -     & -     & 5     & 0.68  & 5     & 0.03  & 5     & 0 \\
          & 4     & -     & -     & 0     & 0.01  & 0     & 0.03  & 0     & 0 \\
          & 5     & -     & -     & 0     & 0     & 0     & 0.02  & 0     & 0 \\
          & 6     & -     & -     & 0     & 0     & 0     & 0.01  & 0     & 0 \\
    \bottomrule
    \end{tabular}%
  \label{tab:addlabel}%
\end{table}%

\begin{table}[htbp]

  \scriptsize
  \centering
  \caption{The comparison between EnumKOpt and three partial MaxSAT solvers on instances from BHOSLIB}
    \begin{tabular}{rC{0.7cm}R{1cm}rrrrrrr}
    \toprule
    \multirow{2}[4]{*}{\textbf{Instance}} & \multirow{2}[4]{*}{\textit{\textbf{k}}} & \multicolumn{2}{c}{\textbf{EnumKOpt}} & \multicolumn{2}{c}{\textbf{Openwho}} & \multicolumn{2}{c}{\textbf{Dist}} & \multicolumn{1}{c}{\textbf{CCLS}} &  \\
\cmidrule{3-10}          &       & \textbf{\#uncov} & \textbf{Time} & \textbf{\#uncov} & \textbf{Time} & \textbf{\#uncov} & \textbf{Time} & \textbf{\#uncov} & \textbf{Time} \\
    \midrule
    \multirow{6}[2]{*}{frb30-15-1} & 1     & -     & -     & -     & -     & 420   & 1.26  & 420   & 0.06 \\
          & 2     & -     & -     & -     & -     & 395   & 56.31 & 399   & 283.88 \\
          & 3     & -     & -     & -     & -     & 370   & 159.99 & 362   & 15.51 \\
          & 4     & -     & -     & -     & -     & 346   & 272.95 & 335   & 34.22 \\
          & 5     & -     & -     & -     & -     & 324   & 102.57 & 307   & 28.75 \\
          & 6     & -     & -     & -     & -     & 300   & 26.85 & 281   & 48.02 \\
    \midrule
    \multicolumn{1}{c}{\multirow{6}[2]{*}{frb30-15-2}} & 1     & -     & -     & -     & -     & 420   & 8.04  & 420   & 0.64 \\
          & 2     & -     & -     & -     & -     & 395   & 47.74 & 399   & 44.59 \\
          & 3     & -     & -     & -     & -     & 370   & 199.81 & 362   & 241.87 \\
          & 4     & -     & -     & -     & -     & 346   & 156.88 & 335   & 104.28 \\
          & 5     & -     & -     & -     & -     & 322   & 286.75 & 308   & 273.68 \\
          & 6     & -     & -     & -     & -     & 300   & 21.35 & 283   & 4.25 \\
    \midrule
    \multicolumn{1}{c}{\multirow{6}[2]{*}{frb30-15-3}} & 1     & -     & -     & -     & -     & 420   & 113.78 & 420   & 2.2 \\
          & 2     & -     & -     & -     & -     & 394   & 192.71 & 399   & 211.42 \\
          & 3     & -     & -     & -     & -     & 371   & 8.03  & 362   & 195.98 \\
          & 4     & -     & -     & -     & -     & 347   & 19.71 & 335   & 119.84 \\
          & 5     & -     & -     & -     & -     & 324   & 139.98 & 309   & 74.73 \\
          & 6     & -     & -     & -     & -     & 300   & 110.5 & 283   & 35.21 \\
    \midrule
    \multicolumn{1}{c}{\multirow{6}[2]{*}{frb30-15-4}} & 1     & -     & -     & -     & -     & 420   & 3.69  & 420   & 0.29 \\
          & 2     & -     & -     & -     & -     & 394   & 280.38 & 398   & 168.84 \\
          & 3     & -     & -     & -     & -     & 370   & 2.81  & 362   & 260.13 \\
          & 4     & -     & -     & -     & -     & 346   & 80.65 & 335   & 45.19 \\
          & 5     & -     & -     & -     & -     & 324   & 56.13 & 308   & 79.48 \\
          & 6     & -     & -     & -     & -     & 398   & 279.9 & 282   & 32.38 \\
    \midrule
    \multicolumn{1}{c}{\multirow{6}[2]{*}{frb30-15-5}} & 1     & -     & -     & -     & -     & 420   & 39.6  & 420   & 0.74 \\
          & 2     & -     & -     & -     & -     & 395   & 68.11 & 398   & 28.87 \\
          & 3     & -     & -     & -     & -     & 370   & 204.62 & 363   & 16.19 \\
          & 4     & -     & -     & -     & -     & 347   & 64.8  & 335   & 65.67 \\
          & 5     & -     & -     & -     & -     & 323   & 92.48 & 308   & 39.54 \\
          & 6     & -     & -     & -     & -     & 299   & 77.83 & 281   & 194.56 \\
    \midrule
    \multicolumn{1}{c}{\multirow{6}[2]{*}{frb40-19-1}} & 2     & -     & -     & -     & -     & 721   & 148.58 & 720   & 10.92 \\
          & 3     & -     & -     & -     & -     & 689   & 13.94 & 695   & 126.36 \\
          & 4     & -     & -     & -     & -     & 655   & 262.24 & 646   & 33.77 \\
          & 5     & -     & -     & -     & -     & 626   & 0.26  & 608   & 219.64 \\
          & 6     & -     & -     & -     & -     & 594   & 79.72 & 575   & 46.84 \\
          & 6     & -     & -     & -     & -     & 564   & 56.26 & 539   & 112.75 \\
    \midrule
    \multicolumn{1}{c}{\multirow{6}[2]{*}{frb40-19-2}} & 1     & -     & -     & -     & -     & 722   & 1.32  & 720   & 171.06 \\
          & 2     & -     & -     & -     & -     & 687   & 270.9 & 696   & 15.23 \\
          & 3     & -     & -     & -     & -     & 656   & 169.92 & 646   & 57.02 \\
          & 4     & -     & -     & -     & -     & 624   & 138.17 & 609   & 182.65 \\
          & 5     & -     & -     & -     & -     & 591   & 169.88 & 573   & 35.85 \\
          & 6     & -     & -     & -     & -     & 563   & 23.17 & 539   & 41.68 \\
    \midrule
    \multicolumn{1}{c}{\multirow{6}[2]{*}{frb40-19-3}} & 1     & -     & -     & -     & -     & 721   & 283.33 & 720   & 8.72 \\
          & 2     & -     & -     & -     & -     & 689   & 39.92 & 695   & 128.3 \\
          & 3     & -     & -     & -     & -     & 656   & 174.28 & 643   & 22.33 \\
          & 4     & -     & -     & -     & -     & 626   & 62.05 & 608   & 258.65 \\
          & 5     & -     & -     & -     & -     & 594   & 109.17 & 570   & 203.09 \\
          & 6     & -     & -     & -     & -     & 564   & 62.48 & 537   & 144.3 \\
    \midrule
    \multirow{6}[2]{*}{frb40-19-4} & 1     & -     & -     & -     & -     & 721   & 263.21 & 720   & 116.17 \\
          & 2     & -     & -     & -     & -     & 688   & 280.11 & 695   & 279.9 \\
          & 3     & -     & -     & -     & -     & 656   & 41.92 & 644   & 131.68 \\
          & 4     & -     & -     & -     & -     & 624   & 75.82 & 606   & 63.13 \\
          & 5     & -     & -     & -     & -     & 594   & 156.85 & 572   & 286.44 \\
          & 6     & -     & -     & -     & -     & 565   & 5.33  & 538   & 57.72 \\
    \midrule
    \multirow{6}[2]{*}{frb40-19-5} & 1     & -     & -     & -     & -     & 722   & 6.05  & 721   & 0.93 \\
          & 2     & -     & -     & -     & -     & 688   & 19.26 & 693   & 238.16 \\
          & 3     & -     & -     & -     & -     & 656   & 118.38 & 645   & 10.66 \\
          & 4     & -     & -     & -     & -     & 624   & 20.03 & 608   & 3.6 \\
          & 5     & -     & -     & -     & -     & 592   & 287.11 & 571   & 106.25 \\
          & 6     & -     & -     & -     & -     & 563   & 20.04 & 536   & 52.28 \\
    \bottomrule
    \end{tabular}%
  \label{tab:addlabel}%
\end{table}%

{\bfseries Part 4:} We perform one experiment in Table 5 to compare the diversified top-$k$ CA solver with three partial MaxSAT solvers on CAs. In Table 5, each instance is denoted by CA$(N,s_1,s_2,\ldots,s_M,t)$, which is defined in Section 5.2. These instances solved by the partial MaxSAT solvers are also first encoded from CAs into diversified top-$k$ partial MaxSAT problem and then converted into partial MaxSAT problem with EE encoding.  From Table 5, we can find that the performance of $k$-CA is roughly worse than the three partial MaxSAT solvers when $k\geq 3$. Thus, it is worth solving diversified top-$k$ CA with converting the problem to diversified top-$k$ partial MaxSAT problem.

\section{Conclusion}
In this paper, we define the diversified top-$k$ partial MaxSAT problem, which generalizes partial MaxSAT problem and enumeration problem. We provide an encoding EE from diversified top-$k$ partial MaxSAT into partial MaxSAT. In order to verify the correctness of EE encoding, we propose an exact algorithm MEMKC. In the experiments, we demonstrate that our approach can be successfully solved the diversified top-$k$ partial MaxSAT problem and this problem can be effectively applied to diversified top-$k$ clique and diversified top-$k$ CA problems.

\begin{table}[htbp]
\setlength{\abovecaptionskip}{-2.pt}
 \setlength{\belowcaptionskip}{5.pt}
  \scriptsize
  \centering
  \caption{The comparison between $k$-CA solver and three partial MaxSAT solvers}
  \scalebox{1.0}{
    \begin{tabular}{cC{0.7cm}rrrrrrrr}
    \toprule
    \multicolumn{1}{r}{\multirow{2}[4]{*}{\textbf{Instance}}} & \multirow{2}[4]{*}{\textit{\textbf{k}}} & \multicolumn{2}{c}{\textbf{$k$-CA}} & \multicolumn{2}{c}{\textbf{Openwbo}} & \multicolumn{2}{c}{\textbf{Dist}} & \multicolumn{2}{c}{\textbf{CCLS}} \\
\cmidrule{3-10}          &       & \textbf{\#uncov} & \textbf{Time} & \textbf{\#uncov} & \textbf{Time} & \textbf{\#uncov} & \textbf{Time} & \textbf{\#uncov} & \textbf{Time} \\
    \midrule
    \multirow{6}[2]{*}{CA(8,2,2,2,2)} & 1     & 9    & 0     & 9    & 0  & 9    & 0  & 9   & 0 \\
          & 2     & 6    & 0     & 6    & 0  & 6    & 0.04  & 6    & 0 \\
          & 3     & 4    & 0     & 3    & 0 & 3    & 0.04  & 3    & 0 \\
          & 4     & 2    & 0     & 0    & 0 & 0    & 0.04  & 0    & 0 \\
          & 5     & 2    & 0     & 0     & 0     & 0    & 0.01  & 0    & 0 \\
          & 6     & 0    & 0     & 0     & 0     & 0    & 0.01  & 0    & 0 \\
    \midrule
    \multirow{6}[2]{*}{CA(16,2,2,2,2,2)} & 1     & 18    & 0     & 18    & 0.01  & 18    & 0.01  & 18    & 0 \\
          & 2     & 12    & 0     & 12    & 0.01  & 12    & 0.01  & 12    & 0 \\
          & 3     & 8    & 0     & 7    & 0.01 & 7    & 0.01  & 7    & 0 \\
          & 4     & 5    & 0     & 2     & 0.03    & 2    & 0.01  & 2    & 0 \\
          & 5     & 2    & 0     & 0     & 0     & 0    & 0.01  & 0    & 0 \\
          & 6     & 0    & 0     & 0    & 0    & 0    & 0.01  & 0    & 0 \\
    \midrule
    \multirow{6}[2]{*}{CA(16,2,2,2,2,3)} & 1     & 28    & 0     & 28    & 0.01   & 28    & 0.01     & 28    & 0 \\
          & 2     & 24    & 0     & 24    & 0.05 & 24    & 0.01  & 24    & 0 \\
          & 3     & 24    & 0     & 20    & 0.84 & 20    & 0.01  & 20    & 0 \\
          & 4     & 24    & 0     & 16     & 4.82    & 16    & 0.02  & 16    & 0 \\
          & 5     & 20     & 0     & 12     & 45.93     & 12    & 0.01  & 12    & 0 \\
          & 6     & 16     & 0     & 8     & 122.59    & 8    & 0.01  & 8    & 0 \\
    \midrule
    \multirow{6}[2]{*}{CA(32,2,2,2,2,2,2)} & 1     & 30    & 0     & 30    & 0  & 30    & 0.01  & 30    & 0 \\
          & 2     & 20    & 0     & 20    & 0.02 & 20    & 0.01  & 20    & 0 \\
          & 3     & 14    & 0     & 12    & 0.14 & 12    & 0.01  & 12    & 0 \\
          & 4     & 8     & 0     & 4     & 3.14     & 4    & 0.01  & 4    & 0 \\
          & 5     & 4     & 0     & 2     & 1046.84     & 2    & 0.01  & 2    & 0 \\
          & 6     & 4     & 0     & 0     & 0     & 0    & 0.01  & 0    & 0 \\
    \midrule
    \multirow{6}[2]{*}{CA(32,2,2,2,2,2,3)} & 1     & 70    & 0     & 70    & 0.10  & 70    & 0.02  & 70    & 0 \\
          & 2     & 60    & 0     & 60    & 1.78 & 60    & 0.02  & 60    & 0 \\
          & 3     & 51     & 0     & 50    & 86.91 & 50    & 23.96  & 50    & 0 \\
          & 4     & 51     & 0     & 40     & 226.6     & 40    & 53.84  & 40    & 0 \\
          & 5     & 42     & 0     & -     & -     & 32    & 0.4  & 32    & 0 \\
          & 6     & 36     & 0     & -     & -     & 24    & 0.02  & 24    & 0 \\
    \bottomrule
    \end{tabular}}%
  \label{tab:addlabel}%
\end{table}%

\end{document}